\pdfoutput=1

\documentclass[11pt]{article}

\usepackage[]{ACL2023}

\usepackage{times}
\usepackage{latexsym}
\usepackage{multirow}
\usepackage{url}
\usepackage{booktabs}
\usepackage[ruled,vlined]{algorithm2e}
\usepackage[T1]{fontenc}

\usepackage[utf8]{inputenc}

\usepackage{microtype}

\usepackage{inconsolata}

\usepackage{tikz,lipsum,lmodern}
\usepackage[most]{tcolorbox}
\usepackage{enumitem}

\title{Augmenting Reddit Posts to Determine Wellness Dimensions impacting Mental Health}

\author{$^{*}$Chandreen Liyanage, $^{**}$Muskan Garg, $^{*}$Vijay Mago, $^{**}$Sunghwan Sohn\\
$^{*}$ Lakehead University,
    Thunder Bay, ON P7B 5E1, Canada\\
  $^{**}$Mayo Clinic, Rochester, MN 55901, USA }

\begin{document}
\maketitle
\begin{abstract}
Amid ongoing health crisis, there is a growing necessity to discern possible signs of Wellness Dimensions (WD)\footnote{The concept of Wellness Dimensions is often used in holistic approaches to health, recognizing that well-being encompasses multiple areas of life.} manifested in self-narrated text. As the distribution of WD on social media data is intrinsically imbalanced, we experiment the generative NLP models for data augmentation to enable further improvement in the pre-screening task of classifying WD. To this end, we propose a simple yet effective data augmentation approach through prompt-based Generative NLP models, and evaluate the ROUGE scores and syntactic/semantic similarity among \textit{existing interpretations} and \textit{augmented data}. Our approach with ChatGPT model surpasses all the other methods and achieves improvement over baselines such as Easy-Data Augmentation and Backtranslation. Introducing data augmentation to generate more training samples and balanced dataset, results in the improved F-score and the Matthew's Correlation Coefficient for upto 13.11\% and 15.95\%, respectively. 
\end{abstract}

\section{Introduction}

\begin{figure}
    \centering
    \includegraphics{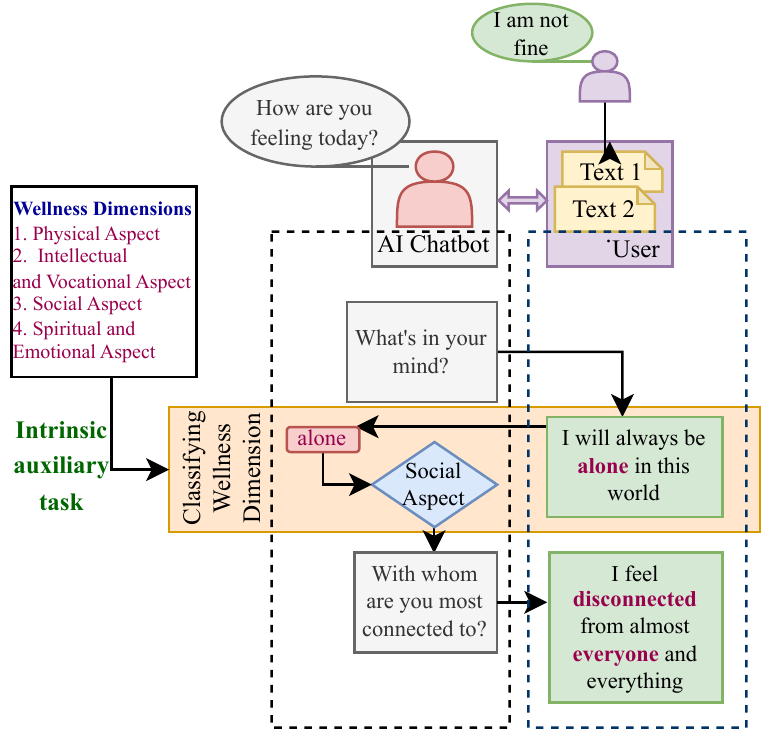}
    \caption{Overview of the task. Generating balanced dataset through data augmentation to facilitate the development of classifiers screening Reddit posts through a lens of Wellness Dimensions.}
    \label{fig:1}
\end{figure}

The social determinants of health (SDOH) refer to various factors present in the surroundings where individuals are born, reside, acquire knowledge, work, engage in leisure activities, practice religion, grow older, impacting a broad range of health-related outcomes, risks and quality-of-life indicators.\footnote{\url{https://health.gov/healthypeople/priority-areas/social-determinants-health}}\footnote{Social Determinants of Health (SDOH) are the social and economic factors that influence an individual's health outcomes.} A rapid expansion of research in SDOH 2030 encourages the social NLP research community to design and develop computational intelligence models for enhancement of an individual's well-being~\cite{bompelli2021social}. In this work, we choose to pre-screen human-writings for biomedical therapy by investigating latent indicators of \textit{wellness dimensions} in Reddit posts (see illustration in Figure~\ref{fig:1}). Wellness dimensions (WD) refer to different aspects of an individual's overall well-being that contribute to their physical, spiritual, social, emotional, intellectual, occupational, environmental, and financial well-being. The disturbed WD, if remains unaddressed, have adverse impact on mental health of an individual. As social media becomes integral part of our daily lives~\cite{wang2020public}, studies in the past suggest that individuals tend to express their thoughts and emotions impacted by one or more wellness dimensions more easily on social media platforms as compared to during in-person sessions with clinical psychologists and mental healthcare~\cite{zhang2023emotion,garg2023mental}. We construct, annotate and observe the original (natural) composition of WD dataset as an imbalanced dataset. In this work, we augment a multi-class dataset on WD to facilitate design and development of NLP models for classifying WD impacting mental health in Reddit posts during mental health screening. Pre-screening filters are helpful in biomedical therapy by facilitating early detection of WD impacting mental health, which if left untreated may cause severe mental disorders. Dunn highlights holistic nature of wellness in 1961 as a \textit{high-level wellness}, denoting a superior level of healthy living~\cite{printz2017applicability}. 

We reduce multiple WD to four key dimensions of well-being based on the frequency and recognition in human writings: Physical Aspect (PA), Intellectual and Vocational Aspect (IVA), Social Aspect (SA), Spiritual and Emotional Aspect (SEA)~\cite{wickramarathne2020review,dillette2021dimensions}. Our major contributions (as illustrated in Fig.~\ref{fig:1}) include (i) the applicability of generative NLP models for domain-specific data augmentation, (ii) examining the diversity among generated and original instances through semantic and syntactic similarity measure, (iii) test and validate the efficacy of data augmentation by investigating classifiers' performance.



\section{Background}
According to Weiss (1975), sociologists put forth a theory that outlines six social needs to prevent loneliness: attachment, social integration, nurturance, reassurance of worth, sense of reliable alliance, and guidance in stressful situations~\cite{weiss1975loneliness}. The Self-Determination Theory (SDT)\footnote{\url{https://en.wikipedia.org/wiki/Self-determination_theory}} highlights the importance of balancing relatedness, competency, and autonomy for intrinsic motivation and genuine self-esteem, which contribute to overall well-being. Neglecting mental disturbance can escalate sub-clinical depression to clinical depression by activating interpersonal risks. This research seeks to examine the origins and outcomes of mental disturbance to mitigate these risks.
\paragraph{Corpus Construction} We present a new dataset with 3,092 instances and 72,813 words to identify wellness dimensions impacting mental disturbance: PA, IVA, SA, and SEA. A senior clinical psychologist, a rehabilitation councilor, and a social NLP researcher framed annotation schemes and perplexity guidelines for text annotation through pre-defined wellness dimensions. Our experts trained three postgraduate students to annotate the data based on predefined dimensions. The annotations were validated using Fleiss' Kappa inter-observer agreement, resulting in a kappa score of 74.39\%. Final annotations were determined through majority voting and expert verification. The experts achieved a kappa score of 87.32\% for the selection of explanatory text spans. Despite slight confusion between PA and SEA, there was a higher agreement for the selection of explanations. To facilitate future research and developments, we publicly release our dataset at Github.\footnote{\url{https://github.com/drmuskangarg/WellnessDimensions}}

\paragraph{Problem Formulation:}
We collect and annotate Reddit data from subreddits \texttt{r/depression} and \texttt{r/suicidewatch} for the task of identifying WD and found imbalanced dataset in its natural composition, suggesting the need of data augmentation. To evaluate the effectiveness of generative NLP models for data augmentation, we frame the task of \textit{augmenting Reddit posts} as a \textit{text generation} problem. We compare and contrast the performance of model trained on data augmented with (i) Generative Pre-trained Transformers (GPT)~\cite{yang2023evaluations}, and (ii) conventional data augmentation approach for NLP such as Easy Data Augmentation (EDA)~\cite{wei2019eda} and Back Translation (BT)~\cite{sennrich2015improving}.



\section{Experiments and Evaluation}
We first generate the data using two-fold measures: (i) traditional data augmentation methods for NLP - EDA and BT, and (ii) prompt-based Generative Pre-trained Transformer models~\cite{ding2023parameter}. We further investigate 
the diversity of the generated samples in comparison to the original samples and fine-tune BERT language model to observe improvements in WD classification, if any. Hinged on the classification results and similarity measures, we select the best model for augmenting WD dataset. 


\subsection{Methods: Data Augmentation}

We use pre-trained generative models\footnote{\url{https://platform.openai.com/docs/models}}~\cite{menggenerating} for this task: (i) ChatGPT models: \textit{gpt-3.5-turbo and gpt-3.5-turbo-0301}, and (ii) other GPT-3 models: \textit{text-curie-001 and text-davinci-003}. The original dataset consists of 3092 samples, with 740, 592, 1139 and 621 records from classes PA, IVA, SA, and SEA respectively. We first split the dataset such that we maximize the number of training samples required for each WD.\footnote{see Appendix~\ref{traintest} for more details.} After augmentation, the training set comprises a total of 4376 records, with an equal distribution of 1094 records per class.

\paragraph{Prompt Design and Parameter Setup:}
As shown in Figure \ref{fig:11}, we design following prompts to produce, a) text similar to the original text (Topic and text), and b) an explanation of newly generated text (text and explanation). 

\begin{figure}
\begin{tcolorbox}[colback=yellow!5!white,colframe=yellow!50!black,
  colbacktitle=yellow!75!black,title=Prompt Designs for GPT models]
   Considering the given topic, generate similar text to the given text.\\
   Topic: \(\ll\)class label as a string\(\gg\) \\
   Text: \(\ll\)original text sentence\(\gg\) \\ 
  
   Similar text:
   \tcblower
      Consider the examples and generate a very short explanation of the given text. \\ \\
   text: \(\ll\)example1-text\(\gg\) \\
   explanation: \(\ll\)example1-explanation\(\gg\) \\
   ...\\
   text: \(\ll\)example5-text\(\gg\) \\
   explanation: \(\ll\)example5-explanation\(\gg\) \\
   text: \(\ll\)original text sentence\(\gg\) \\
   explanation: \(\ll\)original explanation\(\gg\) \\ \\
   text: \(\ll\)augmented text sentence\(\gg\) \\
   explanation:
\end{tcolorbox}
\caption{The prompt designs for generating Text and Explanation.}
    \label{fig:11}
\end{figure}

While designing prompts according to Open-AI prompt design instructions\footnote{\url{https://platform.openai.com/docs/guides/completion/prompt-design}}, we begin with explanation through instructions and examples or both. During the text creation, we only provide instructions to the model. As every text belongs to one of the four pre-defined WD, we provide class name as an input, for example, "Physical Aspect", hypothesizing its contribution towards contextual consciousness required for enhancing similar text generation. Furthermore, the explanation generation is developed as a few-shot learning approach~\cite{brown2020language}, where we provide \textit{five} text-explanation pairs as examples. The selective examples ensure the representation of all four classes and are made static for every call. 
We keep temperature as 0.7 to preserve the creativity/ randomness of generated text.

\begin{figure}[h]
\centering
\includegraphics[width=7.5cm,height=6.5cm]{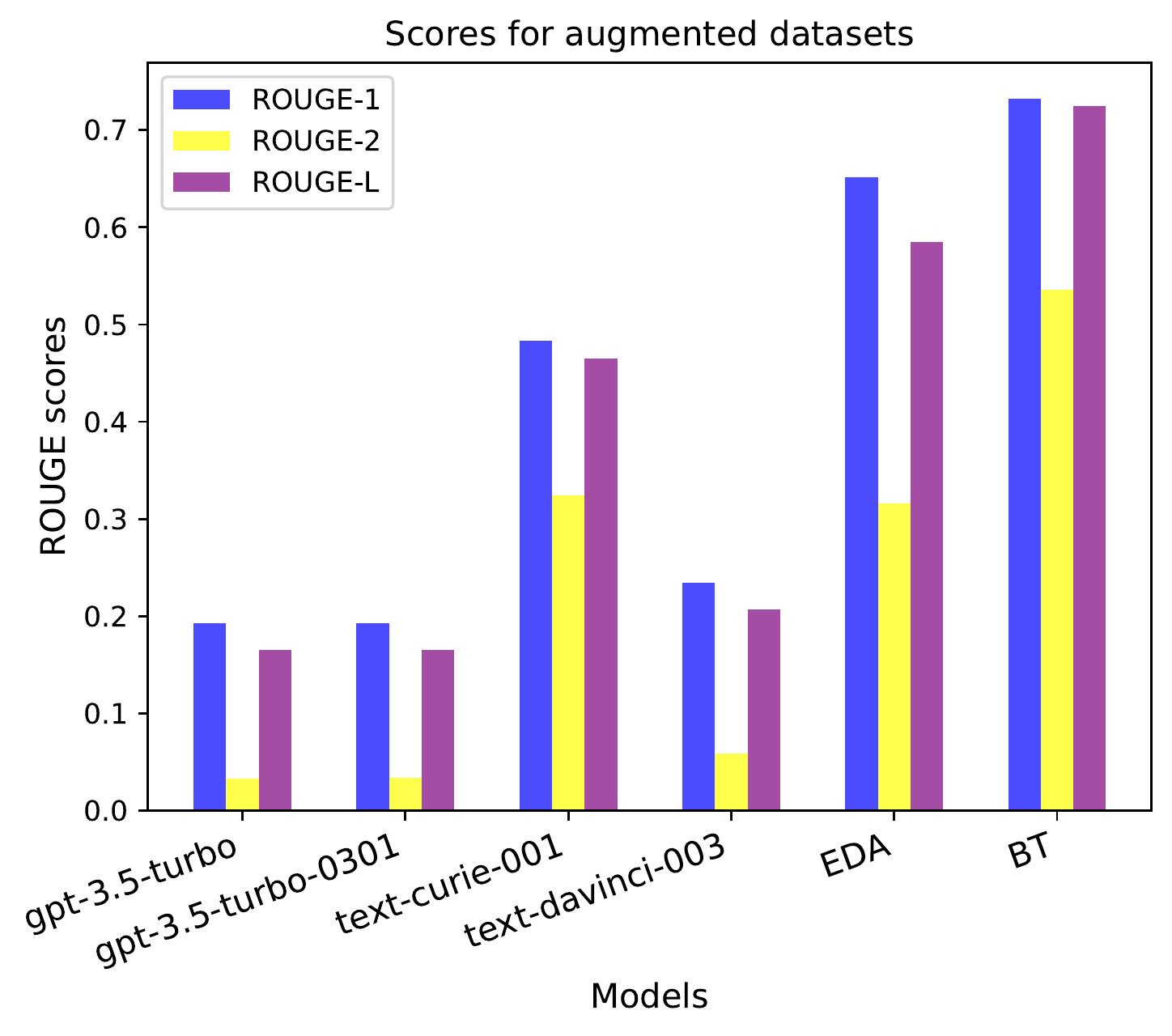}
\caption{We present ROUGE scores for six different augmentation mechanisms leveraging the augmented samples in comparison to the original text, averaged over all the texts generated. }
\label{fig4}
\end{figure}

\begin{figure*}[]
    \includegraphics[width=.49\textwidth]{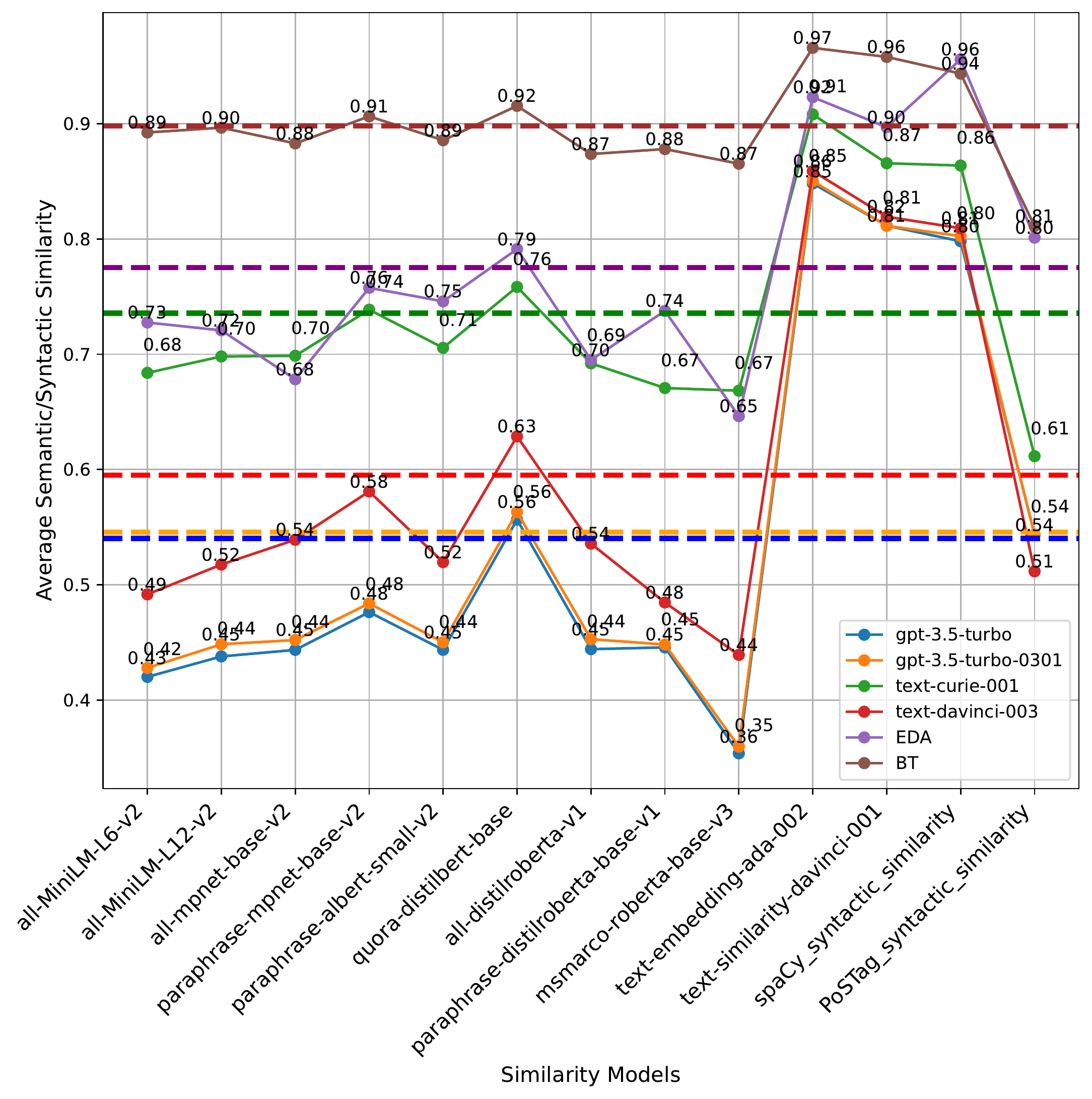}\hfill
    \includegraphics[width=.49\textwidth]{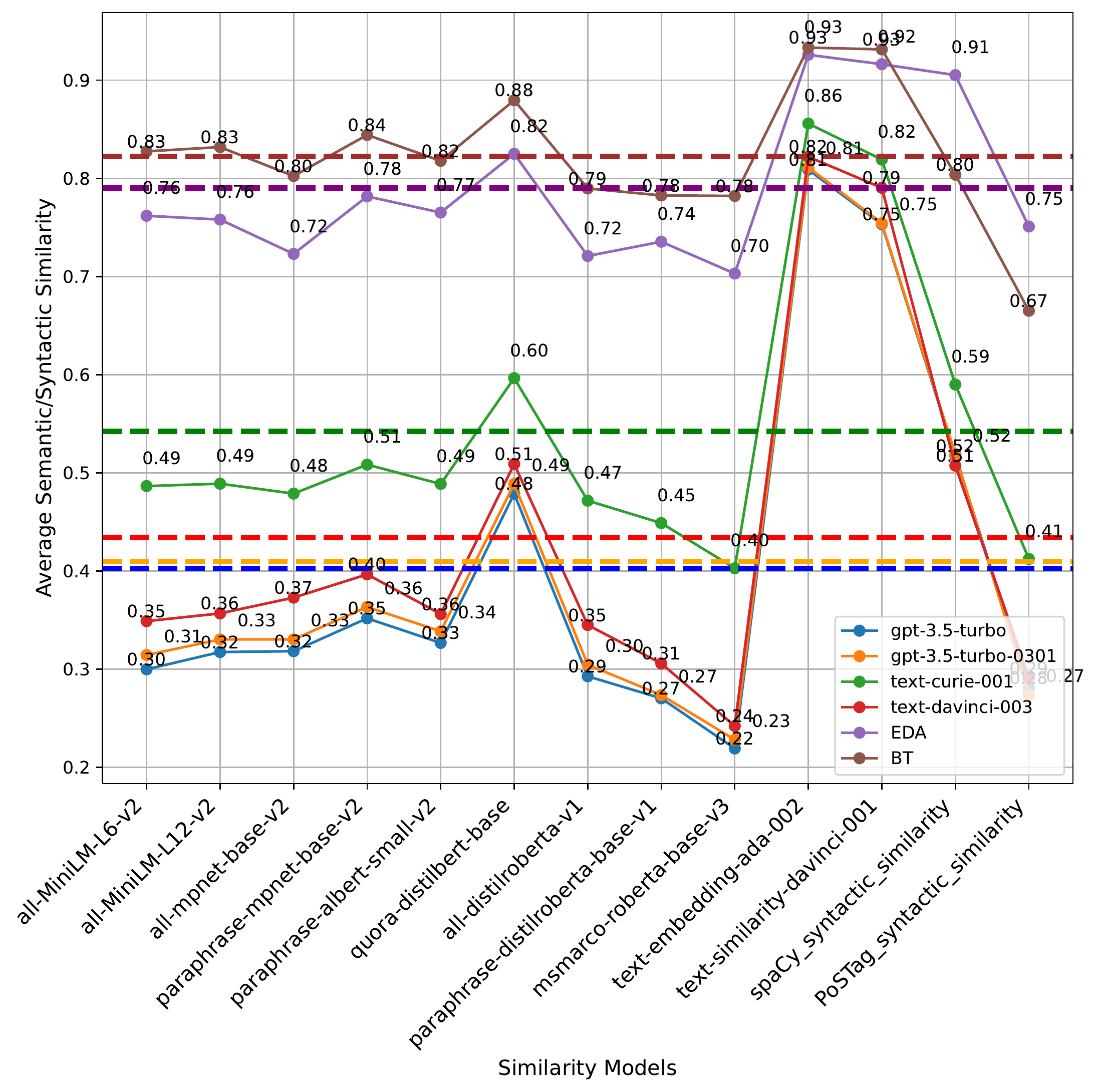}
  
    \caption{(a): Average Textual Similarity among Original and Augmented Text. (b): Average Textual Similarity among Original and Augmented Explanations.}
    \label{fig1}
\end{figure*}


\subsection{Method: Similarity Measures}
First we calculate ROUGE scores <ROUGE-1, ROUGE-2 and ROUGE-L> to examine similarity\footnote{\url{https://pypi.org/project/rouge/}}. Next, for \textit{semantic similarity,} we calculate the embedding for each sentence through eleven pre-trained language models~\cite{ormerod2021predicting, reimers2019sentence}.\footnote{see Appendix~\ref{models} for more details.} The resulting sentence embeddings\footnote{\url{https://platform.openai.com/docs/guides/embeddings/what-are-embeddings}} of each original and augmented data instance were then compared using \textit{cosine similarity}. Lastly, for \textit{syntactic similarity}, we first parsed given sentences into syntactic trees and then mapped them into vector representations using the "en\_core\_web\_md" English pipeline in the spaCy library\footnote{\url{https://spacy.io/models/en}}. Next, these vector representations are used to compute the similarity score between sentences. Furthermore, we compute the set overlap between the Part-of-speech (POS) tag sequences of the original and augmented sentences to determine their similarity\footnote{\url{https://www.nltk.org/api/nltk.tag.pos_tag.html}}. 

\subsection{Classification with BERT}

As the final evaluation, we  build BERT~\cite{kenton2019bert}, a baseline classifier, with 6 augmented datasets and compare its performance with the BERT classifier built over original data. 
We used the training data in WD dataset for finetuning for 10 epochs with a batch size of 32 and a learning rate of 3e-5. To preserve the lengths of texts, we set the max\_length to 256 during  tokenization. We use the validation set (20\% of the training set) and testing set (180 samples) to examine the efficiency and effectiveness of a classifier through F-score and Matthew's Correlation Coefficient (MCC), respectively.

\section{Results and Discussion}
\paragraph{Similarity Analysis:} We report three types of ROUGE scores: ROUGE-1, ROUGE-2 and ROUGE-L 
between the original and augmented text. 
The ChatGPT models show the lowest ROUGE scores, and \textit{gpt-3.5-turbo} versions surpases all other augmentation methods (see Figure~\ref{fig4}). 
We further examine semantic and syntactic similarities through average of all 13 models in Figure~\ref{fig1}(a) and ~\ref{fig1}(b). We observe high diversity and low similarity with GPT based models where ChatGPT based models illustrate the least similarity. However, compared to the other GPT models, text-curie-001 shows a notably higher similarities through all the similarity models.


\paragraph{Classification Performances: }
We obtain the validation accuracy (Val-A), and testing results with precision (T-P), recall (T-R), F-score (T-F), accuracy (T-A) and MCC value (T-MCC) with experimental results for evaluation (see Table~\ref{tab:1}). Even though we keep the testing dataset to be a small chunk of 180 samples, we observe significant difference in the results in training on the \textit{original imbalanced dataset }and \textit{augmented dataset}. The \textit{gpt-3.5-turbo} model over testing dataset outperforms all the baseline models, specifically the original dataset by 2.99\% F-score and 1.47\% Accuracy followed by the second best model: \textit{gpt-3.5-turbo-0301}. Moreover, compared to the best traditional augmentation method (BT), the top ChatGPT model shows 7.81\% improvement in testing accuracy. Notably, the datasets from text-curie-001 and EDA which gained higher similarity values have shown lowest performance on all classification measurements. 

\begin{table}[]
    \centering
       \caption{Improvement in classifiers. M1: gpt-3.5-turbo, M2: gpt-3.5-turbo-0301, M3: text-curie-001, M4: text-davinci-003.}
       \resizebox{0.5\textwidth}{!}{
    \begin{tabular}{c|c|ccc|cc}
    \hline
       \textbf{Type} & \textbf{Val-A} & \textbf{T-P} & \textbf{T-R} & \textbf{T-F} & \textbf{T-A} & \textbf{T-MCC}   \\
         \hline
         Original & 0.427 & 0.65	& 0.63	& 0.61	& 0.63	& 0.514 \\
         \hline
         M1 & 0.504	& \textbf{0.70}	& \textbf{0.69}	& \textbf{0.69}	& \textbf{0.69}	& \textbf{0.596} \\
         M2 & 0.499	& \textbf{0.69}	& \textbf{0.68}	& \textbf{0.67}	& \textbf{0.68}	& \textbf{0.581} \\
         M3 & 0.498	& 0.63	& 0.63	& 0.62	& 0.63	& 0.519 \\
         M4 & 0.502	& 0.66	& 0.67	& 0.66	& 0.67	& 0.559 \\
         EDA & 0.498	& 0.63	& 0.63	& 0.62	& 0.63	& 0.518 \\
         BT & 0.504	& 0.65	& 0.64	& 0.63	& 0.64	& 0.527 \\
         
        \hline
    \end{tabular}
 }
    \label{tab:1}
\end{table}

We further examine the MCC values to determine the effectiveness of the classifier~\cite{boughorbel2017optimal}. MCC values vary between -1 and 1 such that values closer to 0 and 1 suggest increased randomness and perfect prediction towards decision making correspondingly. We found 15.95\% improvement in MCC score when model is trained on augmented training samples with \textit{gpt-3.5-turbo} model. 
 Overall, the augmented text with lowest ROUGE scores, syntactic and semantic similarities showed the highest classification performance on BERT.

Moreover, the following Table \ref{tab:4} compares the class-vise classification performance between the original and M1 (best performed dataset) datasets. We notice a significant improvement in all the measurements of all the classes after augmenting data. Additionally, compared to other classes, the IVA- class with the least number of original samples shows a significantly higher improvement in the number of correctly classified samples.

\begin{table}[]
\small
\centering
\caption{Class-vise classification performance. NS: Number of correctly classified samples, INS: Improvement in NS (in \%), OD: Original dataset, AD: Dataset augmented by M1 method. }
\label{tab:4}
\begin{tabular}{l|l|lll|c|c}
\hline
\textbf{Class} &\textbf{Type} & \textbf{T-P} & \textbf{T-R} & \textbf{T-F} & \textbf{NS}& \textbf{INS} \\ \hline
\multirow{2}{*}{PA} & OD & 0.78 & 0.71 & 0.74 & 32 & \multirow{2}{*}{4.44} \\
 & AD & 0.76 & 0.76 & 0.76 & 34 &  \\
\multirow{2}{*}{IVA} & OD & 0.67 & 0.31 & 0.42 & 14 & \multirow{2}{*}{17.78} \\
 & AD & 0.69 & 0.49 & 0.57 & 22 &  \\
\multirow{2}{*}{SA} & OD & 0.61 & 0.76 & 0.67 & 34 & \multirow{2}{*}{2.22} \\
 & AD & 0.69 & 0.78 & 0.73 & 35 &  \\
\multirow{2}{*}{SEA} & OD & 0.53 & 0.73 & 0.62 & 33 & \multirow{2}{*}{2.22} \\
 & AD & 0.65 & 0.76 & 0.70 & 34 &  \\ \hline
\end{tabular}%

\end{table}


\section{Conclusion and Future Scope}
In this work, we augment the Reddit posts for a four-class classification problem of determining Wellness Dimensions impacting mental health. The GPT models are outperforming in terms of generating diverse text by preserving the context of the corresponding original text. In future, we plan to experiment with different parameter settings and prompts for generating datasets and develop improved classifiers to determine WD in a well balanced dataset. Furthermore, we will evaluate the classification performance of short explanation text we generated in this dataset.

    \label{tab:my_label}
\section*{Ethics and Broader Impact}
The data used in this study is obtained from Reddit, a platform designed for anonymous posting, and the user IDs have been anonymized. Furthermore, all sample posts displayed in this study have been obfuscated, paraphrased, and anonymized to protect user privacy and prevent any misuse. As annotation is subjective in nature, we acknowledge that there may be some biases present in our gold-labeled data and the distribution of labels in Wellness Dimensions dataset. We urge researchers to be mindful of the potential risks associated with WD dataset based on personal textual information. To prevent this, human intervention by a moderator is necessary. We acknowledge that we do not release user's metadata and the augmented samples further increase the privacy. The dataset and the source code required to replicate the baseline results can be accessed at Github.\footnote{\url{https://github.com/drmuskangarg/WellnessDimensions}}

\section*{Acknowledgement}

We express our gratitude to Veena Krishnan, a senior clinical psychologist, and Ruchi Joshi, a rehabilitation counselor, for their unwavering support throughout the project. This project was partially supported by NIH R01 AG068007. This project is funded by NSERC Discovery Grant (RGPIN-2017-05377), held by Vijay Mago, Department of Computer Science, Lakehead University, Canada.

\bibliography{anthology,custom}
\bibliographystyle{acl_natbib}

\appendix

\begin{table}[h]
\caption{Language models used to evaluate generate sentence embedding}
\label{tab:3}
\resizebox{\columnwidth}{!}{%
\begin{tabular}{|l|l|lll}
\cline{1-2}
\textbf{Base Model} & \textbf{Version}                                        & \textbf{} &  &  \\ \cline{1-2}
BERT                & {\color[HTML]{212121} all-MiniLM-L6-v2}                 &           &  &  \\ \cline{1-2}
BERT                & {\color[HTML]{212121} all-MiniLM-L12-v2}                &           &  &  \\ \cline{1-2}
MPNet               & {\color[HTML]{212121} all-mpnet-base-v2}                &           &  &  \\ \cline{1-2}
MPNet               & {\color[HTML]{212121} paraphrase-mpnet-base-v2}         &           &  &  \\ \cline{1-2}
Albert              & {\color[HTML]{212121} paraphrase-albert-small-v2}       &           &  &  \\ \cline{1-2}
DistilBERT          & {\color[HTML]{212121} quora-distilbert-base}            &           &  &  \\ \cline{1-2}
DistilRoberta       & {\color[HTML]{212121} all-distilroberta-v1}             &           &  &  \\ \cline{1-2}
DistilRoberta       & {\color[HTML]{212121} paraphrase-distilroberta-base-v1} &           &  &  \\ \cline{1-2}
Roberta             & {\color[HTML]{212121} msmarco-roberta-base-v3}          &           &  &  \\ \cline{1-2}
GPT-3               & {\color[HTML]{212121} text-embedding-ada-002}           &           &  &  \\ \cline{1-2}
GPT-3               & {\color[HTML]{212121} text-similarity-davinci-001}      &           &  &  \\ \cline{1-2}
\end{tabular}%
}
\end{table}
\section{Training and Testing Split}
\label{traintest}
Consider the data containing $D$ documents representing a collection of Reddit posts \{$D=d_1, d_2,..., d_n$\} where $n=3092$. For each document $d_i$, there exist a tuple representing $<E_i, C_i>$ where $E_i$ is text-span/ explanation and $C_i$ is the aspect class for $i^{th}$ instance. Thus, the original WD dataset consists of three columns for 3092 samples: $<D_i, E_i, C_i>$. The aspect class $C_i \in \alpha $ where $ \alpha = $[PA, IVA, SA, SEA] and the composition of original dataset contains imbalanced distribution of aspect classes (see Table~\ref{tab:2}). The number of samples for every class $\alpha[j]$ where $1 \le j \le 4$ suggests the need of data augmentation to facilitate development of NLP models over balanced dataset. To this end, we propose the algorithm - \textit{Required annotation count} to decide the number of samples that needs to generated for each WD. 
Given an input $\alpha$ as a list of the number of text samples for different WD (PA, IVA, SA, SEA) where PA, IVA, SA, SEA defines the count of instances for each class. 

As such, our goal is to achieve a balanced dataset by obtaining $1094$ samples for each WD, resulting in $1094 * 4 = 4376$ data samples. We observe that all the samples for IVA class must be augmented while no augmentation is required for SA class.

\begin{algorithm}[]
\label{algo:1}
\SetAlgoLined
\KwResult{Return $AS$=[]} \textcolor{blue}{\tcp{AS: augmented sample}}

\textbf{Input}: $\alpha:$ [PA, IVA, SA, SEA]

\textbf{Set}: $\beta$= [], R, Red=[], RC=[] 

\textbf{Set}: $min_{value} := min(\alpha)$ \tcp{\textcolor{blue}{Get the record count of minority class}}

\For{j in count($\alpha$)}{
        $\beta[j] := max(\alpha) - \alpha[j]$ 
        
        \textcolor{blue}{\tcc{For each class, get the record count difference from majority class}}
     }
\textcolor{blue}{\tcp{Estimate the size of the test set }}
$R= min_{value} - max(\beta[j])$ 

\For{j in count($\alpha$)}{
       \textcolor{blue}{\tcc{For each class, calculate the Reduction Percentage (percentage of reduction after separating the testing set)}}
        
        $Red[j]= \frac{R}{\alpha[j]} * 100$

        \textcolor{blue}{\tcc{For each class, calculate the Reduced Composition (number of training records before augmentation)}}
        
        $RC[j]= \alpha[j] - Red[j]$
     }

\textcolor{blue}{\tcc{Get the maximum number of records per class for augmentation}}
$max_{RC}= max(RC)$ 

\For{j in count($\alpha$)}{
    \textcolor{blue}{\tcc{Augment each class up to the maximum record count}}
        $AS[j]= max_{RC} - RC[j]$ \\
     }
\textcolor{blue}{\tcp{return the augmented dataset}}
\textbf{return $AS$}
\caption{Required Augmentation Count}
\end{algorithm}

\section{Semantic Similarity Models}
\label{models}
We use 11 different semantic similarity models as shown in Table~\ref{tab:3}. Sentence Transformers are a set of state-of-the-art language models implemented in Python for generating text embeddings. The two different GPT-3 models used for this task accessed the embeddings API endpoint.

\begin{table}[]
\small
    \centering
    \begin{tabular}{c|ccc|c|c}
    \toprule[1.5pt]
        WD & $\alpha$ & Red & RC & AS & Tot.\\
        \midrule
         PA &  740 & 6.0 & 695 & 399 & 1094\\
         IVA & 592 & 7.6 & 547 & 547 & 1094\\
         SA &  1139 & 4.0 & 1094 & 0 & 1094 \\
         SEA & 621 & 7.2 & 576 & 518 & 1094 \\ 
         \bottomrule[1.5pt]
    \end{tabular}
    \caption{The statistics of original composition ($\alpha$), the reduction percentage (Red), reduced composition (RC), the number of augmented samples (AS) and total number of samples (Tot.)}
    \label{tab:2}
\end{table}


\end{document}